\documentclass[10pt,twocolumn,letterpaper]{article}

\usepackage{cvpr}
\usepackage{times}
\usepackage{epsfig}
\usepackage{graphicx}
\usepackage{amsmath}
\usepackage{amssymb}
\usepackage{enumitem}

\cvprfinalcopy 

\def\cvprPaperID{****} 

\begin{document}

\title{Window-Object Relationship Guided Representation Learning \\for Generic Object Detections}

\author{ Xingyu ZENG \qquad Wanli OUYANG \qquad Xiaogang WANG\\
The Chinese University of Hong Kong\\
{\tt\small \{xyzeng,wlouyang,xgwang\}@ee.cuhk.edu.hk}
}

\maketitle

\begin{abstract}
In existing works that learn representation for object detection, the relationship between a candidate window and the ground truth bounding box of an object is simplified by thresholding their overlap. This paper shows  information loss in this simplification and picks up the relative location/size information discarded by thresholding. We propose a representation learning pipeline to use the relationship as supervision for improving the learned representation in object detection. 
Such relationship is not limited to object of the target category, but also includes surrounding objects of other categories.
We show that image regions with multiple contexts and multiple rotations are effective in capturing such relationship during the representation learning process and in handling the semantic and visual variation caused by different window-object configurations.
Experimental results show that the representation learned by our approach can improve the object detection accuracy by $6.4\%$ in mean average precision (mAP) on ILSVRC2014 \cite{ILSVRC15}. On the challenging ILSVRC2014 test dataset \cite{ILSVRC15}, 48.6\% mAP is achieved by our single model and it is the best among published results. 
On PASCAL VOC, it outperforms the state-of-the-art result of Fast RCNN \cite{girshick2015fast} by 3.3\% in absolute mAP. 
\end{abstract}

\section{Introduction}
Object detection is the task of finding the bounding boxes of objects from images. It is challenging due to variations in illumination, texture, color, size, aspect ratio, deformation, background clutter, and occlusion. In order to handle these variations, good features for robustly representing the discriminative information of objects are critical. Initially, researchers employed manually designed features \cite{Lowe:SIFT, Dalal:HOG, ojala2002multiresolution}. Recent works  \cite{Krizhevsky:ImageNetCNN, Le:DBNLargeScale, simonyan2014very,Lin2014,chatfield2014return} have demonstrated the power of learning features with deep neural networks from large-scale data. It advances the state-of-the-art of object detection substantially \cite{girshick2014rich, sermanet2013overfeat, zou2014generic, he2014spatial, szegedy2014going, ouyang2014deepid}. 

\begin{figure}[!ht]
\begin{center}
   \includegraphics[width=0.82\linewidth]{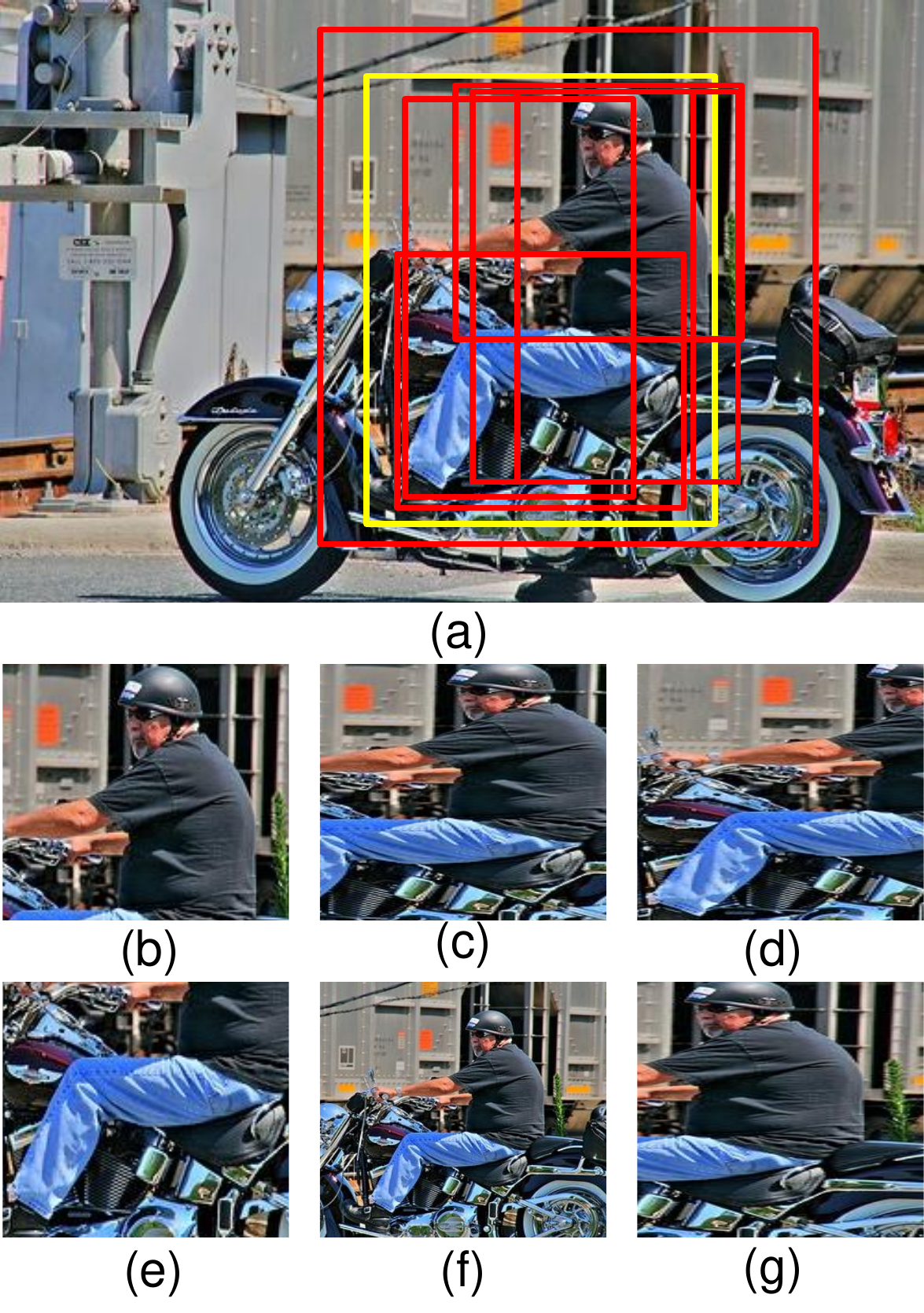}
\end{center}
   \caption{ Examples of candidate windows for detecting persons in an image. (a) Image with a man riding a motorbike. The yellow rectangle denotes the ground truth bounding box of a person. The red rectangles denote several candidate windows whose overlaps with the ground truth are larger than $0.5$. Existing methods assign all these candidate windows to class ``person'' when learning feature representation, despite the large variation of visual cues and different semantic regions covered. (b) and (e) are candidate windows containing the upper body or the legs of the person. (c) and (f) are candidate windows with a smaller or larger size than the ground truth. (d) and (g) are candidate windows containing the left/right  body of a person. }
\label{fig:figure1}
\vspace{-1pc}
\end{figure}

Representation learning for object detection was considered as a multi-class problem \cite{girshick2014rich,donahue2014decaf}, in which a candidate window is classified as containing an object of category $c$ or background, decided by thresholding the overlap between the candidate window and the ground truth bounding box.

In this paper, we show that representation learning for object detection is beyond a multi-class problem. The relationship between the candidate window and the ground truth bounding box of the object, which is called the window-object relationship in this paper, provides rich information to guide representation learning for object detection. However, such information is lost in existing representation learning frameworks which largely simplify the window-object relationship by threhsolding the overlap. Some examples of person detection are shown in Figure \ref{fig:figure1}. The candidate windows in Figure \ref{fig:figure1}(b)-(g) may contain the upper body (a) or the legs (e) of a person, the left (c) or right (f) body of the person, and may have a smaller (c) or larger (d) size than the ground truth. They are all labeled as the same class ``person'' in existing representation learning frameworks for object detection, because their overlaps with the ground truth bounding box are all above $0.5$. However their visual content and semantic meanings have significant difference. If the deep neural network is required to classify all these candidate windows into the same class, it is easy for the model to get confused and it becomes difficult to learn representation capturing semantically meaningful visual patterns, since the supervision is weak. Such ambiguity can be can resolved by using the window-object relationship as supervision during training, which well reflect all types of variations mentioned above. Being aware of these variations in supervision, it is easier for the model to disentangle these variation factors in the learned representations.



The contributions of this work are summarized below. 
First, we propose a representation learning pipeline by using the window-object relationship as supervision so that the learned features are more sensitive to locations and sizes of objects. By distinguishing and predicting window-object relationship, the learned representation captures more semantically meaningful visual patterns of candidate windows on objects. Experimental results show that the representation learned by our approach can improve mAP of object detection by $6.4\%$ on ILSVRC2014. 

Second, two objective functions are designed to encode the window-object relationship. Since the window-object relationship is complex, our experiments show that direction prediction on the relative translation and scale variation  in a similar way as bounding box regression does not improve representation learning. Instead, under each object category, we cluster candidate windows into subclasses according to the window-object relationship. Both visual cues and window-object relationship of candidate windows in the same subclass have less variations. Given the cropped image region of a candidate window as input, the deep neural network predicts the subclass as well as the relative translation and scale variation under the subclass during representation learning. Different subclasses employ different regressors to estimate the relative translation and scale variation.  

\begin{figure}[t]
\begin{center}
   \includegraphics[width=1.05\linewidth]{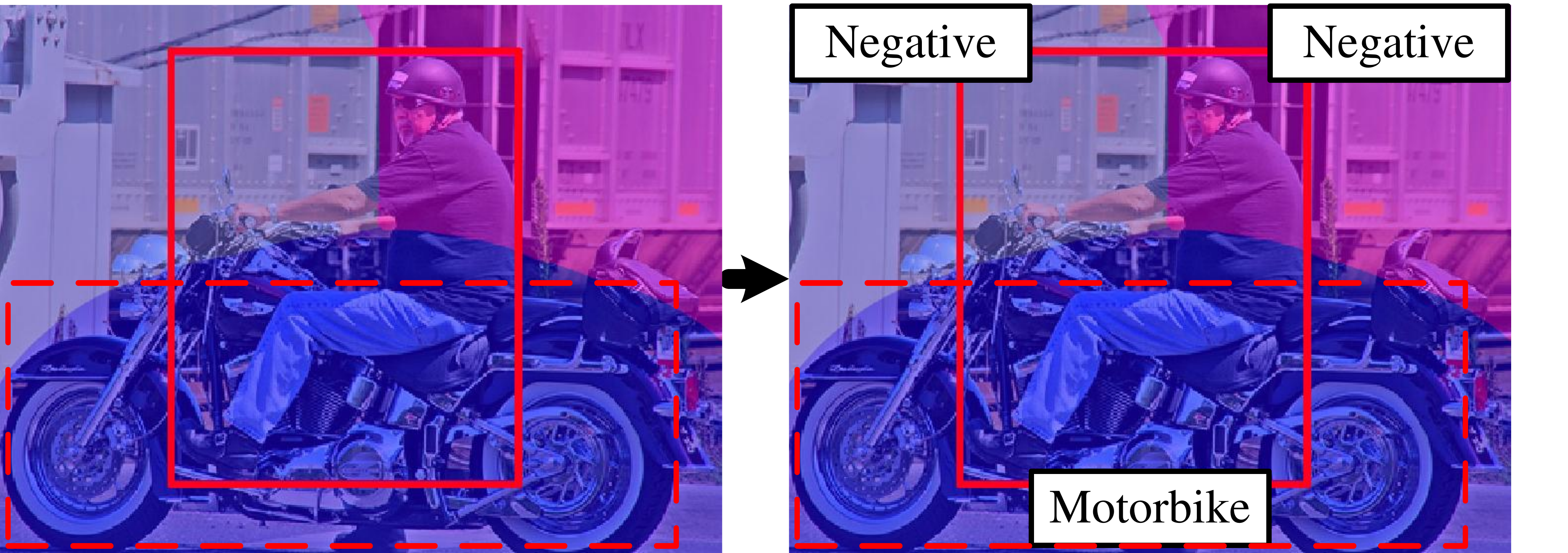}
\end{center}
   \caption{ Examples of candidate windows for detecting persons in an image. The neighbor is separated into three regions, and labels indicating other neighbor objects, i.e., motorbike in the dash rectangle,  are utilized to help feature representation learning process. Best viewed in color.}
\label{fig:figure2}
\vspace{-1pc}
\end{figure}

Third, the idea is also extended to model the relationship between a candidate window and objects of other classes in its neighborhood, given the cropped image region of the candidate window. An illustration is shown in Figure \ref{fig:figure2}.  The learned feature representation can make such prediction because it captures the pose information indicating existence of neighbor objects from the cropped image region (e.g. the pose of the person in Figure \ref{fig:figure2} indicates that he rides on a motorbike) and the image region may include parts of neighbor objects. All these disturbing factors explained in Figure \ref{fig:figure1} and \ref{fig:figure2} are nonlinearly coupled in the image region and deteriorate the detection accuracy. With window-object relationship as supervision, they are disentangled in the learned feature representation and can be better removed in the later fine-turning stage or by a SVM classifier. 

Fourth, we show that the window-object relationship can be better modeled by taking image regions with multiple contexts and multiple rotations as input, which includes multiple types of contextual information. This is different from commonly used multi-scale deep models, which take the same image region of different resolutions as input. Compared with the baseline, the multi-context and multi-rotation input improves the mAP by $2.2\%$ on ILSVRC2014. By adding the supervision of window-object relationship on multi-context and multi-rotation, the mAP was further improved by $4.2\%$ on ILSVRC2014. 


\section {Relative Work}
RCNN \cite{girshick2014rich} is a widely used object detection pipeline based on CNN \cite{simonyan2014very,szegedy2014going,ouyang2014deepid}. It first pre-trains the representation by classifying $1.2$ million images from ImageNet into $1,000$ categories and then fine-tunes it by classifying object detection bounding boxes on the target detection dataset. People improved RCNN by proposing better structures of CNN \cite{szegedy2014going,simonyan2014very}. Ouyang \textit{et al.} \cite{ouyang2014deepid} improved pre-training by classifying the bounding boxes of the images from ImageNet instead of the whole images. All these works posed representation learning as a multi-class problem without effort on exploring window-object relationship.


A group of works tried to solve detection with regression \cite{szegedy2013deep,szegedy2015object,yoo2015attentionnet}. Given the whole image as input, Szegedy \textit{et al.} \cite{szegedy2013deep} used DNN to regress the binary masks of an object bounding box and its subboxes. Szegedy \textit{et al.} \cite{szegedy2015object} used CNN to directly predict  the coordinates of object bounding boxes. AttentionNet \cite{yoo2015attentionnet} initially treated the whole image as a bounding box, and iteratively refined it. They quantified the way of adjusting the bounding box into several directions, and made decision at each step. Since the locations and sizes of objects in images have large variations, direct prediction is challenging. Although some promising results were obtained on PASCAL VOC, these works have not reported state-of-the-art result on ImageNet yet, which includes a much larger number of object categories and test images. AttentionNet required training separate networks for different categories and is not scalable. It only reported the result of one category (i.e. ``human'') on PASCAL VOC and its average precision is lower than ours by $2\%$, while our learned representation is shared by a large number of categories. Different from these approaches, we explore window-object relationship to improve representation learning, while our test pipeline is similar as RCNN. Moreover, we observe that directly predicting the locations and sizes of candidate windows does not improve representation learning, since the window-object relationship is complex. Supervision needs to be carefully designed. 


In RCNN, bounding box regression was used as the last step to refine the locations of candidate windows. However, it was not used to learn feature representation. The recently proposed Fast RCNN \cite{girshick2015fast} jointly predicted object categories and locations of candidate windows as multi-task learning. $0.8\%$ meanAP improvement is observed on PASCAL 07 dataset. However, this multi-task learning only improves meanAP by $0.2\%$ point in the ILSVRC2014.


In this paper, multi-context and multi-rotation input is used. The related work \cite{gidaris2015object} cropped multiple subregions as the input of CNN. Besides enriching the representation, our motivation of employing multi-context and multi-rotation input is to make CNN less confused about the relationship between candidate windows and objects. Details will be given in Section \ref{sec:multiscal_rotation}.


\begin{figure*}[t]
\begin{center}
   \includegraphics[width=0.92\linewidth]{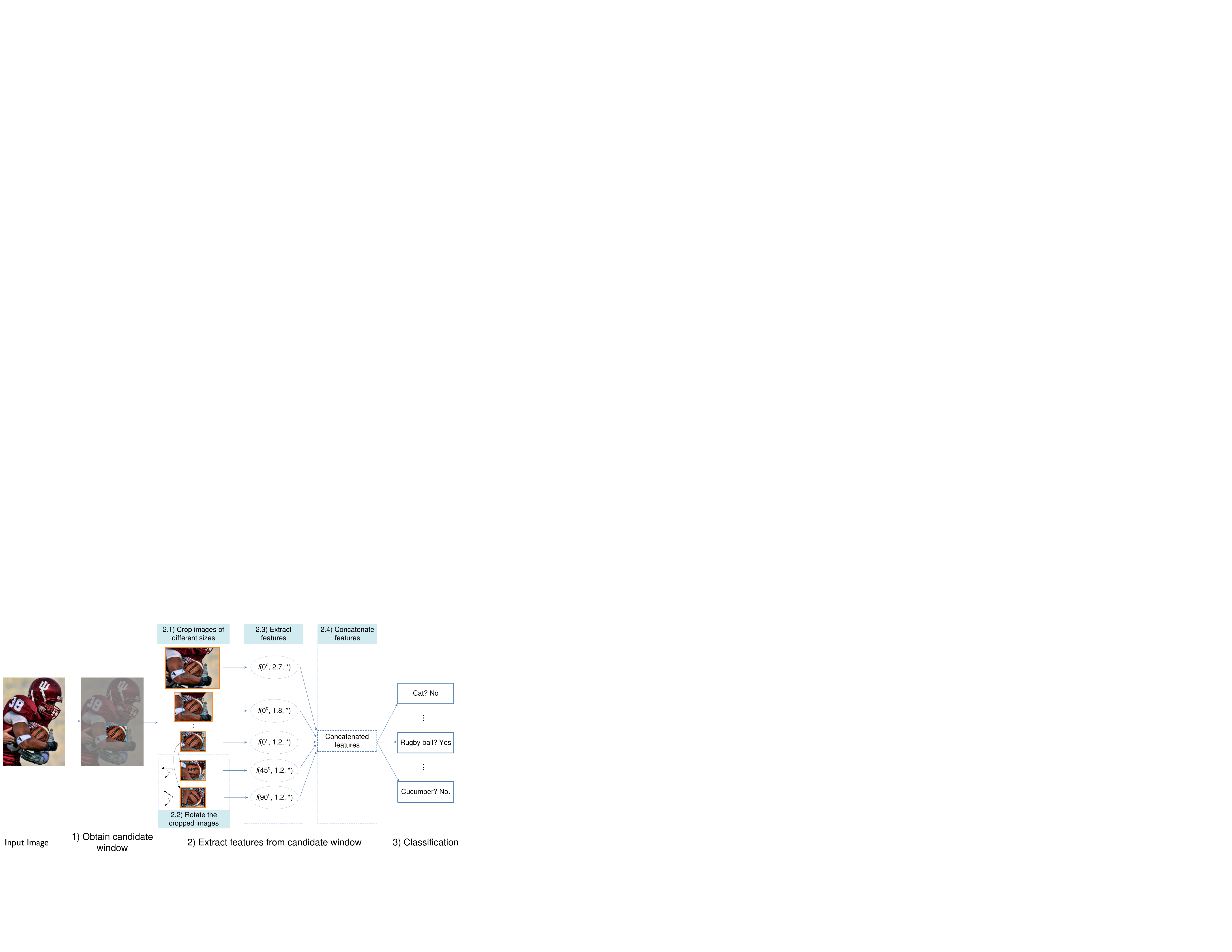}
\end{center}
   \caption{Object detection at the test stage.  1) For a given candidate window, images cropped with 2.1) different sizes $\lambda$ and 2.2) rotation degrees $r$ are warped into the same size. 2.3) For the cropped image of a given rotation degree and scale $(r, \lambda)$, a CNN $f(r, \lambda, *)$ is used for extracting  features. 2.4) Features for multiple scales and rotations are concatenated and 3) used for classification. }
\label{fig:detpipeline}
\vspace{-1pc}
\end{figure*}
\section{Method}
In order to provide readers with a clear picture of the whole framework, we first explain the object detection pipeline at the test stage. The major contributions comes from representation learning, whose details are provided in Section \ref{Sec:TrainPipeline} - Section 
\ref{sec:N_finetune}.

\subsection{Object detection at the testing stage}
As in Fig. \ref{fig:detpipeline}, the object detection pipeline is as follows:
\begin{enumerate}[leftmargin=12pt,noitemsep,nolistsep, label=\bfseries \arabic*)]
\item  Selective search in \cite{Smeulders:SelectiveSearch} is adopted to obtain candidate windows. 
\item  A candidate window is used to extract features as below:
\begin{enumerate}[leftmargin=24pt,noitemsep,nolistsep, label=\bfseries 2.\arabic*)]
\item  For a candidate window $\mathbf{b}_s = (x,y,W,H)$ with size $(W, H)$ and center $(x,y)$, crop images $I(\lambda, \mathbf{b}_s)$ with sizes  $(\lambda W, \lambda H), \lambda \in \Lambda$ and center $(x,y)$. The cropped images and the candidate window have the same center location $(x,y)$. $\lambda$ is the scale of  a contextual region.
The choice of the scale set ${\Lambda}$ is detailed in Section \ref{sec:multiscal_rotation}.
\item  Rotate the cropped image by degrees $r \in \mathbb{R}$ and pad it with surrounding context to obtain  $I(r, \lambda,  \mathbf{b}_s)$,
 $\mathbb{R}=\{0^{\circ}, 45^{\circ}, 90^{\circ}\}$.
\item The cropped images $I(r,\lambda,  \mathbf{b}_s)$ with different sizes and rotations are warped into the same size and treated as the input of CNN for extracting their features, i.e. $\mathbf{f}_{r,\lambda} = f(r, \lambda, I(r,\lambda,  \mathbf{b}_s))$ where $f(r, \lambda, *)$ denotes the CNN for extracting features from $ I(r,\lambda,  \mathbf{b}_s)$, $\mathbf{f}_{r,\lambda}$ denotes the vector of features extracted for rotation $r$ and scale $\lambda$. For the candidate window $ \mathbf{b}_s$, there are six cropped images  $ I(r,\lambda,  \mathbf{b}_s)$ with $(r,\lambda)$ being $(0^\circ, 0.8), (0^\circ, 1.2), (45^\circ, 1.2), (90^\circ, 1.2),  (0^\circ, 1.8)$, and $(0^\circ, 2.7)$ in our experiment.  In the experiments, the structure of CNN is chosen as GoogleNet \cite{szegedy2014going} for different settings of $(r, \lambda)$. And there are six branches of GoogleNets for the six settings of $(r, \lambda)$. The learned parameters for the six branches of GoogleNets are different.
\item The extracted features are then concatenated into $\mathbf{F}=concat_{ (r,  \lambda)}\{ \mathbf{f}_{r,\lambda} \}$, where $concat$ is the operation for concatenating features into a vector.
\end{enumerate}
\item  Extracted features are used by $C$ binary-class SVM to classify each candidate window. The  score of each SVM measures the confidence on the candidate window containing a specific object class.
\end{enumerate}
The steps are similar to RCNN \cite{girshick2014rich} except for multi-context and multi-rotation input. Our major novelties come from how to train the feature extractor $f$ in Step 2.3 of Fig. \ref{fig:detpipeline}.

\subsection{Representation learning pipeline}
\label{Sec:TrainPipeline}

Our proposed pipeline is as follows and shown in Fig. \ref{fig:trainpipeline}. 
\begin{enumerate}[leftmargin=12pt,noitemsep,nolistsep, label=\bfseries \alph*)]
\item  \label{pipeline:step1} Pretrain CNN using the ImageNet 1000-class classification and localization data.
\item  \label{pipeline:step2} Use the CNN trained in the previous step for initialization. Train the CNN by estimating the window-object relationship. Details are given in Section \ref{sec:locationPretrain}. 
\item  \label{pipeline:step3} Use the CNN trained in the previous step for initialization.  Train the CNN by estimating the window-multi-objects relationship. Details are given in Section \ref{sec:window_object_multi_objects}.
\item   \label{pipeline:step4} Use the CNN trained in the previous step for initialization. Train the CNN for $C$+1-classification problem. $C$ is the number of object classes, plus 1 for background. $C=20$ for PASCAL VOC and $C=200$ for ILSVRC2014.  Details are given in Section \ref{sec:N_finetune}.
\end{enumerate}
Since the pipeline above is used for learning feature representation, except for the difference in the output layer, the network structures are the same for all the training steps above. And the responses of the last CNN layer before the output layer are treated as feature representation.

%

\subsubsection{Window-object relationship label preparation}

\subsection{Learning the window-object relationship} 
\label{sec:locationPretrain}

\begin{figure}[t]
\begin{center}
   \includegraphics[width=0.83\linewidth]{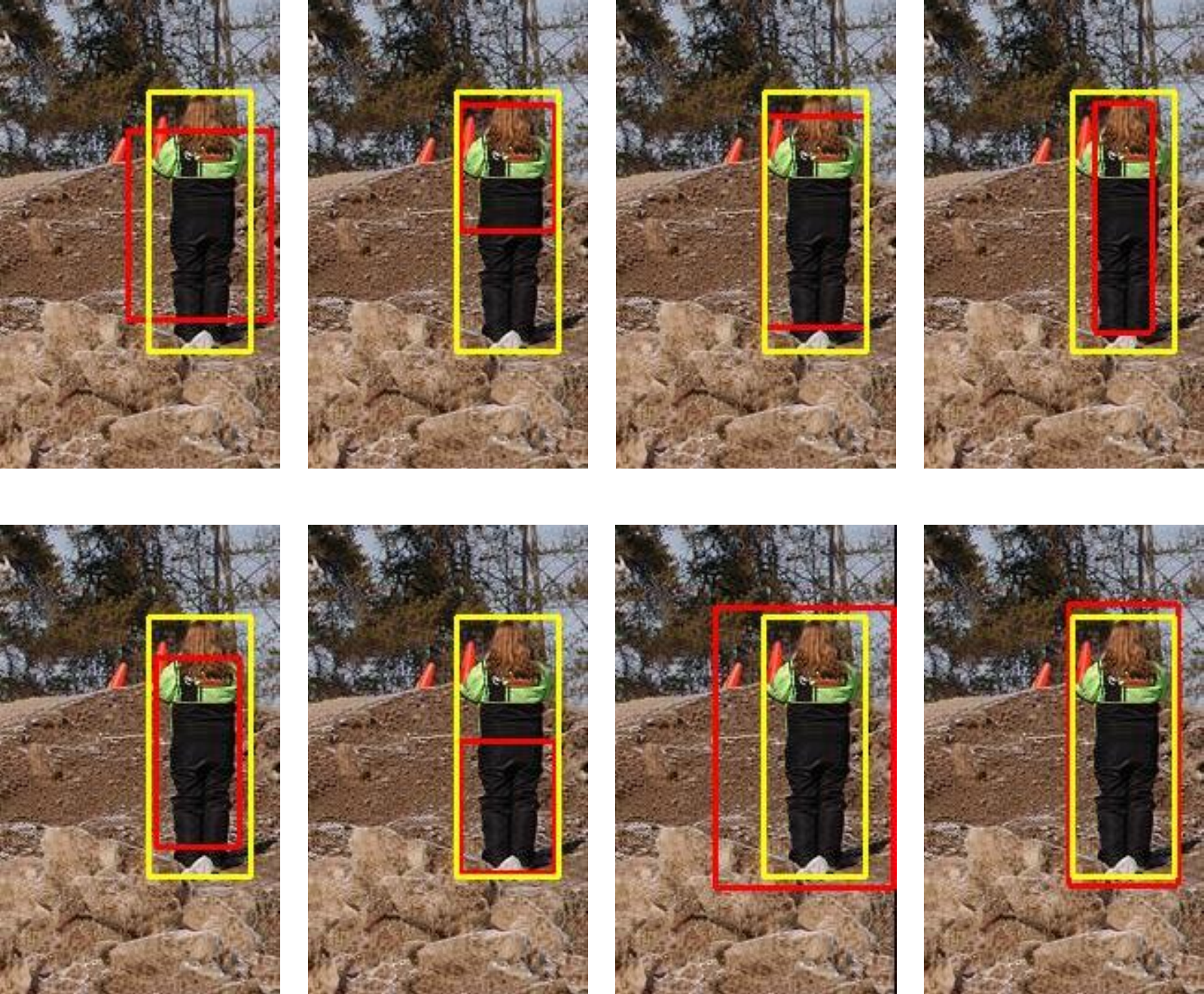}
\end{center}
   \caption{Examples of window-object relationship clusters obtained. Yellow rectangles denotes ground truth. Red rectangle denotes a window-object relationship cluster (only showing the average of the clustered candidate windows). Best viewed in color.}
\label{fig:cluster}
\vspace{-1.3pc}
\end{figure}

The  idea is to have  CNN distinguish  candidate windows containing different parts of the same object or having different sizes. For example, a candidate window containing the upper body of a  person and another one containing the legs were classified as the same category (i.e. ``person'') in existing works, but are considered as different configurations of window-object relationship in our approach.

\begin{figure*}[!ht]
\begin{center}
   \includegraphics[width=0.9\linewidth]{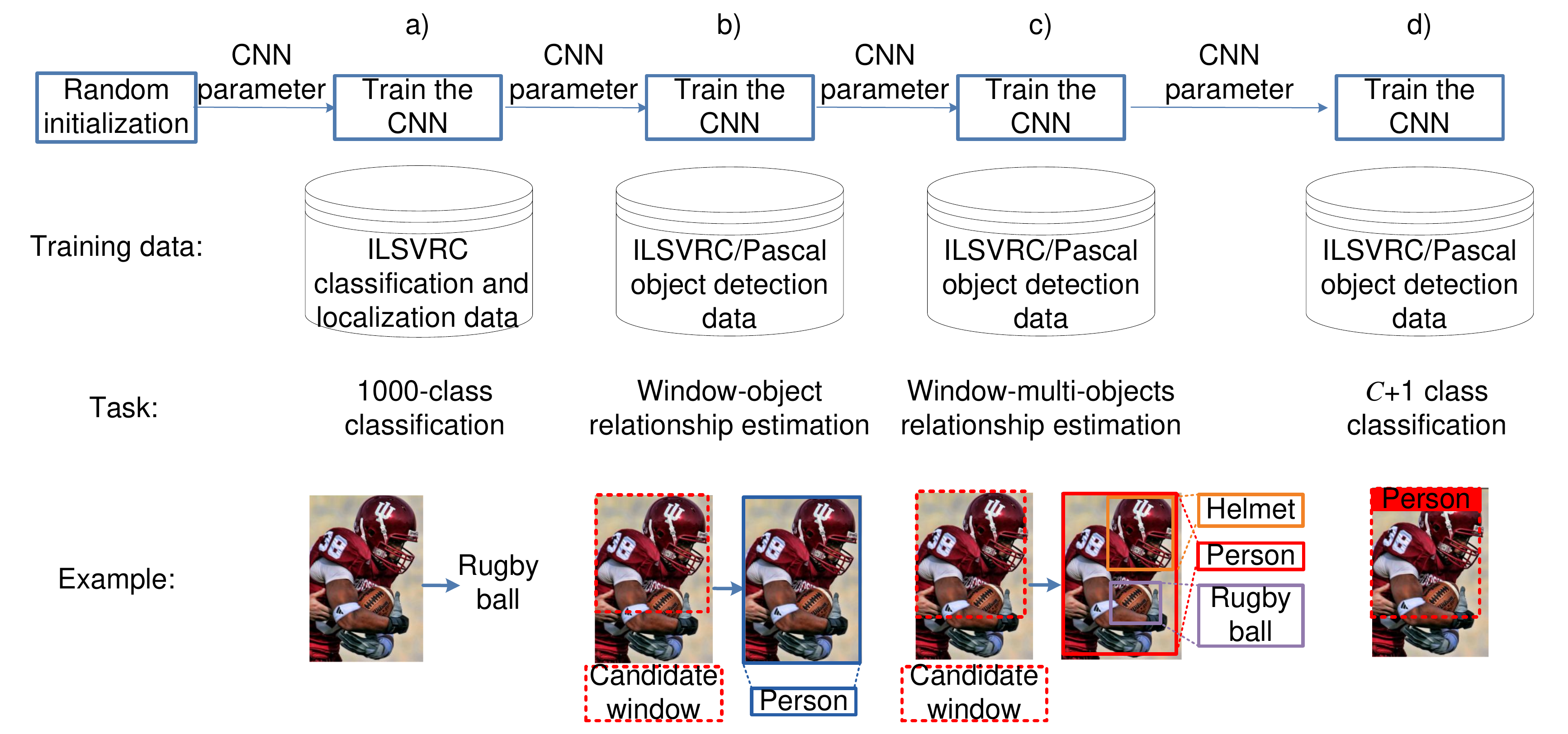}
\end{center}
   \caption{Overview of the representation learning pipeline. Best view in color.}
\label{fig:trainpipeline}
\vspace{-1pc}
\end{figure*}


To distinguish candidate windows of the same object class, we cluster  training samples in each class into subsets with similar relative locations. Denote by $\mathbf{b}_{i,s}=(x_{i,s}, y_{i,s}, W_{i,s}, H_{i,s})$ the  $i$-th candidate window at the training stage with center $(x_{i,s}, y_{i,s})$ and size $(W_{i,s}, H_{i,s})$. Its ground-truth bounding box is denoted by $\mathbf{b}_{i,g}=(x_{i,g}, y_{i,g}, W_{i,g}, H_{i,g})$. Candidate windows at the training stage are from  selective search \cite{Smeulders:SelectiveSearch} and ground truth bounding boxes.
The relative location and size between the candidate window and the ground-truth bounding box (normalized by the size of candidate window) are: 
\begin{align}
\mathbf{l}_{i, loc}=\big{[}& (x_{i,s}-x_{i,g})/W_{i,s}, &(y_{i,s}-y_{i,g})/H_{i,s}, \\
&log(W_{i,s}/W_{i,g}), &log(H_{i,s}/H_{i,g}) \big{]}.
\label{equ:cluster}
\end{align}
The relative location and size above are used for describing the window-object relationship.
With features $\mathbf{l}_{i, loc}$, affinity propagation(AP) \cite{frey2007clustering} is used to group candidate windows with similar window-object relationship into $N$ clusters. Denote the cluster label for the $i$th candidate window by $n_{i} \in \{1, \ldots, N\}$. Fig. \ref{fig:cluster} shows some clustering results where each cluster corresponds to a specific visual pattern and relative location setting. 
%
In window-object relationship prediction, the labels for a given candidate window are the relative location and size $\mathbf{l}_{i, loc}$, and the cluster label $l_{i, cls}$.

\subsubsection{Loss function of window-object relationship}
\label{sec:location_specific_labels_loss}
With the CNN parameters obtained from  Step \ref{pipeline:step1} in Section \ref{sec:locationPretrain} as initialization, we continue to train the CNN by predicting  window-object relationship. The CNN's  1000-way classification layer in Step \ref{pipeline:step1} is replaced by two fully connected (fc) layers. The layer that predicts the location and size for cluster $n$, denoted by $\{\tilde{\mathbf{l}}_{i, loc, n}|  n=1, \ldots, N\}$, is called the location prediction layer. The other layer that predicts the cluster label $\tilde{n}_{i}$ for the $i$th candidate window is called the cluster prediction layer. Both layers use the last feature extraction layer of the CNN as input. The output dimension of the location prediction layer is $4N$ and the output dimension of the layer that outputs $\tilde{n}_{i}$ is $N$. Softmax is used for the cluster prediction layer.
%
The following loss on  window-object relationship is used:
\begin{align}
L &= L_{cls} + L_{loc}, \label{eq:Loss1}\\
L_{cls}&=-\sum_{i}{n_{i}\log(\tilde{n}_{i})},  \nonumber \\
L_{loc}&=\sum_{i}{\| \tilde{\mathbf{l}}_{i, loc, n_i} -\mathbf{l}_{i, loc} \|_{2}^{2}}. \nonumber
\end{align}
$L_{cls}$ is the loss on predicting the window-object relationship cluster, $L_{loc}$ is the loss on predicting the relative location and size. With the loss in (\ref{eq:Loss1}), CNN is required to distinguish different window-object relationship clusters and to recognize where the actual relative location is. For the $i$th candidate window, the CNN outputs $N$ location prediction vectors, i.e. $\tilde{\mathbf{l}}_{i, loc, n}$ for $n=1, \ldots, N$ so that the CNN learn the location prediction for different clusters separately. For the $i$th candidate window, only the cluster $n_i$ is used for supervising the location prediction.
Therefore, different window-object relationship clusters have their own parameters learned separately by CNN to prediction location. For example, the location bias for the cluster with window-above-object relationship can be different from the bias for the cluster with window-below-object relationship.
Since the relative locations in the same cluster have less variation, this divide-and-conquer strategy makes prediction easier.
With the loss function defined,  CNN parameters are learned by BP and stochastic gradient descent.

\subsubsection{Multi-context  and multi-rotation}
\label{sec:multiscal_rotation}
\begin{figure}[t]
\begin{center}
\includegraphics[width=0.9\linewidth]{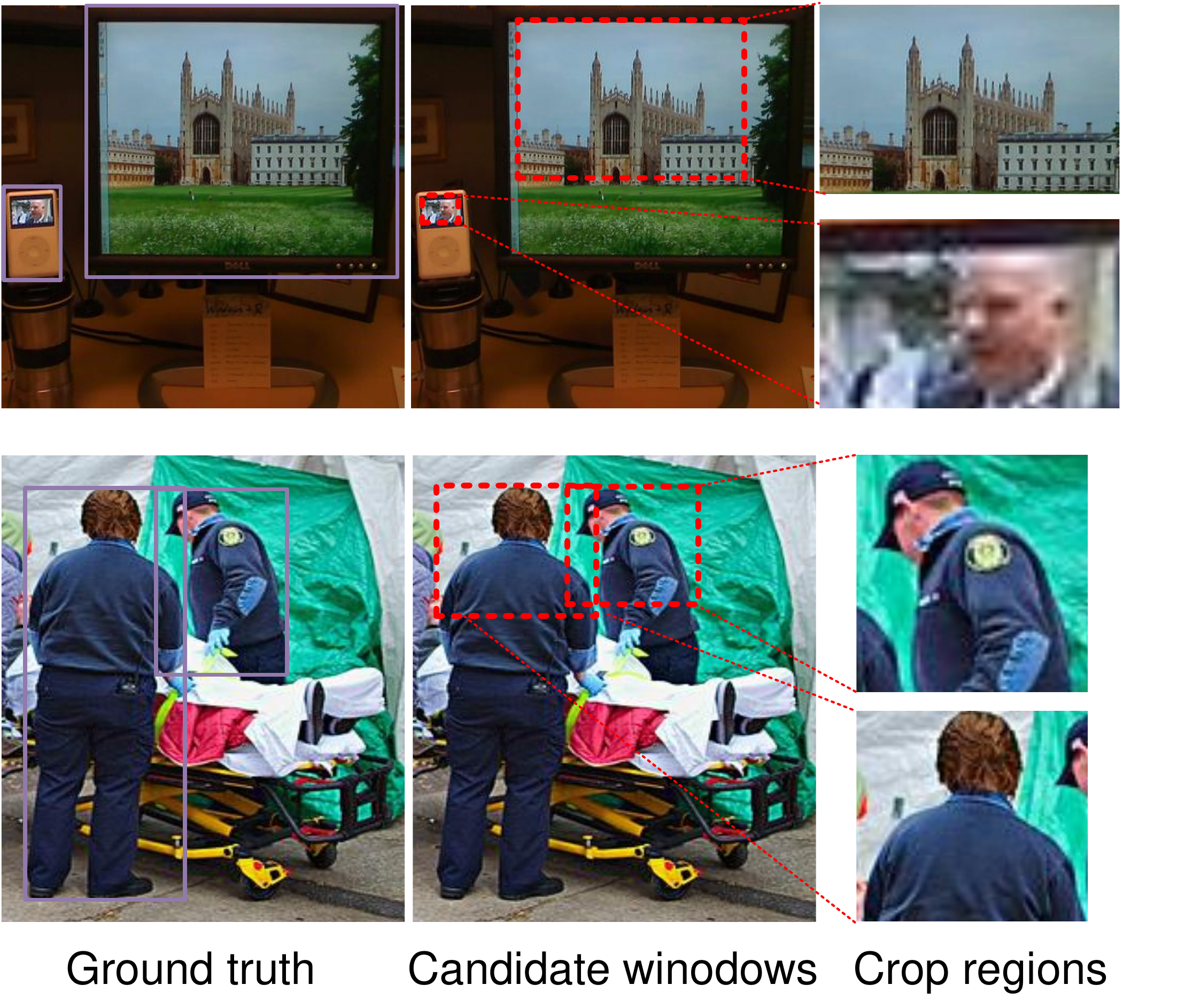}
\end{center}
\vspace{-0.5pc}
\caption{
It is hard to tell  object class (top row) or where the ground truth bounding box is (bottom row) if the only the image region within the candidate window is provided. Best viewed in color.}
\label{fig:gt_candidate}
\vspace{-1pc}
\end{figure}

When the location and size of a candidate window is different from that of the ground truth bounding box, the candidate window only have partial visual content of the object. The limited view  results in difficulty for  CNN to figure out the visual difference between object classes. For example, it is hard to tell whether it is an ipod or a monitor if one can only see the screen, but it becomes much easier if the whole object and its contextual region is provided, as shown in Fig. \ref{fig:gt_candidate} (top row). When occlusion happens, the ground truth bounding boxes may contain different amount of object parts and thus have different sizes. Without a region larger than the ground truth as input, it is confusing for CNN to decide the bounding box size. In Fig. \ref{fig:gt_candidate} (bottom row), the ground truth box for a standing unoccluded person should cover more parts of human body than the one with legs occluded. When the image region cropped from a candidate window only covers the upper body of this person, it is difficult to predict whether the person's legs are occluded or not.  When predicting the relative location between the candidate window and the ground truth, CNN should output a smaller box if occluded, but a larger box otherwise.  CNN can handle this difficulty when the input contains a larger region than the ground truth. On the other hand, if the region is much larger than the object, the resolution of the object may not be high enough after normalizing the cropped image region to a standard size as the input of CNN.

To handle the problems above, we use multiple scales of contextual regions as the input for CNN. The feature learning procedure still focuses on predicting the window-object relationship. We use 4 scales for cropping images, 0.8, 1.2, 1.8, 2.7, which are linear in log scale. 1.2 is the only scale chosen in \cite{girshick2014rich} and is set as default value in many existing works. In the supplementary material, we prove that the cropped image with scale 2.7 is sufficient to cover most of the ground-truth region when the overlap between the window and the object is greater than 0.5. Even if the overlap between the candidate window and ground truth bounding box is 0.37, the cropped image with scale 2.7 can cover more than 50\% of the ground truth region.  1.8 is obtained by linear interpolation between 1.2 and 2.7 in log scale.
0.8 is chosen because some candidate windows can be larger than the ground truth bounding box, as shown by the first image in Fig. \ref{fig:cluster}. A cropped image with a smaller scale can help these windows to fit the actual scale of the object.

Object rotation results in drastic appearance variation. Rotation is adopted together with multiple scales for the cropped images to make the network more robust to appearance variations.

Regarding the relative locations labels are used in this training step, we choose to not merge training samples with different scales and rotations together and adopt multiple networks with each network is trained with one kind of training samples. All those networks share the same network structure but different parameters. 

\subsection{Window-multi-object relationship prediction}
\label{sec:window_object_multi_objects}
The previous training step does not consider the coexistence of multiple object instances in the same image, which happens frequently and forms layout configurations. For the example in Fig. \ref{fig:trainpipeline}, the person have a helmet on his head and a rugby ball in his arms. To further enrich the feature representation, we extend the training procedure by predicting the window-multi-objects relationship. 

The window-multi-objects relationship can be formulated in answering three basic questions, whether other instances exist in neighborhood, where they are and what they are. 
We start with the relative location and size  $\mathbf{l}_{i, loc}$ defined in (\ref{equ:cluster})  to describe the pairwise relationship between a candidate window and multiple objects. The  $\mathbf{l}_{i, loc}$ for all ground truth bounding boxes are used as features to obtain  $K$ clusters, which are used for describing the window-multi-objects layout. Given a candidate window, its surrounding ground truth objects are assigned to their closest clusters and $K$ labels are obtained with each label standing for what kind of object in the corresponding cluster. CNN has $K$ classification layers and each layer is a multi-class classifier for its corresponding configuration of window-object relationship. Therefore, CNN is required to predict the probability of what object exists in each location cluster. 

We keep the loss function discussed in \ref{sec:location_specific_labels_loss} and add $K$ cross entropy loss terms. The weights of all losses are set to be 1. In this step, three kinds of labels are applied to each training sample, 1) the window-object cluster label, 2) the relative location between the window and the object, 3) and $K$ labels to represent window-multi-object relationship. 

\subsection{Finetuning for $C$+1-classification}
\label{sec:N_finetune}
Since the ultimate goal is to detect $C$ classes of objects, we use CNN obtained in the previous step as initialization and continue to train it for the $C+1$ classification problem. Cross entropy loss function is used.  As discussed in section \ref{sec:multiscal_rotation}, several scales and rotation degrees are adopted, the networks for different rotations or scales are jointly learned. The features $\mathbf{f}_{r,\lambda}$ extracted from CNN for all scales $\lambda$ and rotations $r$ are concatenated into a vector of features for the $C+1$-class classification problem. 
Once features are learned, we fix the CNN parameters and learn 200 class-specific linear SVMs for object detection as in \cite{girshick2014rich}. 
Be reminded that although multiple concepts, such as window-object relationship, clusters of locations and sizes, other object instances and window-multi-object-relationship, are proposed in the training stage, their goal is to improve the learning of feature representation $f$ in Fig. \ref{fig:detpipeline} and none of them appears in test. 


\section{Experimental results}
\subsection{Experimental setting}
The implementation of our framework adopts GoogleNet \cite{szegedy2014going} as CNN structure. 1000-class pretraining is based on the ILSVRC2014 classification and localization dataset. The learned representation is evaluated on the two datasets below. Most evaluation on  component analysis of our training pipeline is conducted on ILSVRC2014 since it is much larger in scale and contains more object categories. The overall results and comparison with the state-of-the-art are finally reported on both datasets. 

\textbf{The ILSVRC2014 object detection dataset} contains 200 object categories and is split into three subsets, i.e. train, validation and test data. The validation subset is split into val1 and val2 in \cite{girshick2014rich}. We follow the same setting. In the training step \ref{pipeline:step4}, we use both train and val1 subsets, but in the training step \ref{pipeline:step2} and \ref{pipeline:step3}, we use only val1 subset. Because many positive samples are not labeled in the train subset and it may bring in label noise for window-object relationship and window-multi-object relationship.

\textbf{The PASCAL VOC2007 dataset} contains 20 object categories. Following the most commonly used approach in \cite{girshick2014rich}, we finetune the network with the trainval set and evaluate the performance on the test set. 

\subsection{Component analysis on the training pipeline}
\subsubsection{Comparison with baselines}
\label{sssec:baseline}
In order to evaluate the effectiveness of our major innovations, several baselines are compared on ILSVRC2014 and the results are summarized in Table \ref{table:baseline}. (1) RCNN choosing GoogLeNet as the CNN structure. It is equivalent to removing step \textcolor{red}{\textbf{b)}} and \textcolor{red}{\textbf{c)}} in our training pipeline and only taking the single context of scale $1.2$ without rotation as input in test. (2) Since our method concatenates features from seven GoogLeNets, one may question the improvement comes from model averaging. This baseline randomly initializes seven GoogLeNets, concatenate their features to train SVM and follow the RCNN pipeline. (3) Take multi-context and multi-rotation input (i.e. using the test pipeline in Fig. \ref{fig:detpipeline}) without supervision of window-object relationship (i.e. removing step \textcolor{red}{\textbf{b)}} and \textcolor{red}{\textbf{c)}} from our training pipeline). (4) Our pipeline with single context of scale 1.2 and without rotation. (5) Our pipeline excluding window-multi-objects relationship (i.e. step \textcolor{red}{\textbf{c)}}) in training. (6) Our complete pipeline.

The result shows that increasing model complexity by model averaging bring marginal improvement. Multi-context and multi-rotation input improves the RCNN baseline by $2.2\%$ and adding  supervision of window-object relationship to it further obtains the gain of $4.2\%$ in mAP, which is significant. Window-multi-objects relationship contributes $0.5\%$ gain in mAP. 
The gain of adding supervision of window-object relationship to multi-context and multi-rotation input is larger than that added to a single-context input.
It indicates that multi-context and multi-rotation input helps CNN better predict window-object relationship. 

\begin{table}[t]
\centering
\small
\begin{tabular}{c|cccccc}
\hline
 & (1) & (2) & (3) & (4) & (5) & (6) \\
  \hline
mean AP (\%) &39.9& 41.7 & 42.1 &42.9 & 45.8 &46.3\\
  \hline
median AP (\%) &39.7 &41.0& 42.5 & 42.0 &45.7 &45.9\\
  \hline
\end{tabular}
\vspace{0.1in}
\caption{Comparison with several baselines on ILSVRC2014 val2. Find descriptions of methods (1)-(4) in Section \ref{sssec:baseline}. }
\label{table:baseline}
\vspace{-1pc}
\end{table}

\begin{table}[t]
\centering
\small
\begin{tabular}{c|c|c|c}
 \hline
 Approach & Step \ref{pipeline:step1}+\ref{pipeline:step4}  &  Step \ref{pipeline:step1}+\ref{pipeline:step2}+\ref{pipeline:step4} &   Step \ref{pipeline:step1}+\ref{pipeline:step2}+\ref{pipeline:step4}  \\
&& w.o. cluster& w. cluster\\
  \hline
Mean AP (\%) & 39.9  & 40.1 & 41.1\\
  \hline
Median AP (\%) & 39.7  & 39.9 & 41.9 \\
  \hline
\end{tabular}
\vspace{0.1in}
\caption{Effectiveness of clustering window-object relationship  for learning CNN on ILSVRC2014 val2. Step \ref{pipeline:step1}+\ref{pipeline:step4}  corresponds to the training pipeline of RCNN.   Step \ref{pipeline:step1}+\ref{pipeline:step2}+\ref{pipeline:step4} w.o. cluster corresponds to learning window-object relationship without clustering.  Step \ref{pipeline:step1}+\ref{pipeline:step2}+\ref{pipeline:step4} w. cluster corresponds to our approach in clustering window-object relationship. }
\label{table:location_specific_labels}
\vspace{-1pc}
\end{table}

\begin{table*}[!ht]
\centering
\begin{tabular}{c|c|c|c|c|c|c|c}
 \hline
 Scale & 1.2  &  1.2+0.8 & 1.2+1.8& 1.2+2.7 & 0.8+1.2+1.8  & 0.8+1.2+1.8+2.7 & shared\\
  \hline
Mean AP (\%) & 41.1 & 42.1  & 43.6 & 44.2 & 44.7  &45.5& 42.1\\
  \hline
Median AP (\%) & 41.9&  40.9 & 43.1 & 44.1 & 44.8   & 45.9 & 42.5\\
  \hline
\end{tabular}
\vspace{0.1in}
\caption{Influence of using multiple scales on ILSVRC2014 val2. Shared denotes the approach with network parameters shared for four scales (0.8+1.2+1.8+2.7). }
\label{table:scale}
\end{table*}

\begin{table*}[!ht]
\centering
\begin{tabular}{c|c|c|c|c}
 \hline
scale & 1.2 & 1.2 & 0.8+1.2+1.8+2.7 & 0.8+1.2+1.8+2.7 \\
 \hline
 rotation degree & 0$^\circ$ & 0$^\circ$+45$^\circ$+90$^\circ$ & 0$^\circ$ & 0$^\circ$+45$^\circ$+90$^\circ$ \\
  \hline
val2 meanAP (\%) & 41.1 & 43.1 &45.5 & 45.8\\
  \hline
val2 median AP (\%) & 41.9& 42.2   &45.9 & 45.7\\
  \hline
\end{tabular}
\vspace{0.1in}
\caption{Influence of using multiple scale and multiple rotations on ILSVRC2014 val2. Anti-cloclwise rotation is used. }
\label{table:rotation}
\end{table*}

\begin{table*}[!ht]
\centering
\begin{tabular}{ccccccccc}
\hline
 approach&
Flair \cite{vandeSandeCVPR2014} & RCNN\cite{girshick2014rich}  &Berkeley Vision & UvA-Euvision & DeepInsight & DeepID-Net & GoogleNet & ours \\
 \hline
val2(sgl) & n/a  & 31.0  & 33.4 & n/a & 40.1 & 38.5 &38.8 &  49.1  \\
 \hline
val2(avg) & n/a  & n/a  & n/a & n/a & 42 &  40.9 &44.5&   \\
 \hline
test(sgl)& n/a & 31.4 & 34.5 & 35.4 & 40.2 & 37.7&38.0 &  48.6 \\
 \hline
test(avg) & 22.6  & n/a  & n/a & n/a & 40.5 & 40.7  &43.9 &  \\
 \hline
\end{tabular}
\vspace{0.1in}
\caption{Object detection mAP (\%) on ILSVRC2014  for top ranked participants in ILSVRC 2014 with single model(sgl) and averaged model(avg). }
\label{table:ilsvrc14_det}
\end{table*}

\begin{table*}[!ht]
\centering
{\small
\begin{tabular}{p{1.2pc}c|p{0.6pc}p{0.6pc}p{0.6pc}p{0.6pc}p{0.9pc}p{0.6pc}p{0.6pc}p{0.6pc}p{0.6pc}p{0.6pc}p{0.6pc}p{0.6pc}p{0.8pc}p{1pc}p{1.2pc}p{0.6pc}p{0.9pc}p{0.6pc}p{0.6pc}p{1pc}|c}
 \hline
  & data & aero  & bike &  bird & boat &  bottle &  bus & car  & cat  & chair  & cow  & table  & dog &  horse & mbike  & person  & plant & sheep  & softa & train & tv & mAP \\
  \hline
 RCNN& 07 & 73.4 & 77.0  &  63.4 & 45.4 & 44.6  & 75.1 & 78.1 & 79.8  & 40.5 &73.7  & 62.2 & 79.4 & 78.1 & 73.1 & 64.2 & 35.6 & 66.8 & 67.2&70.4&71.1& 66.0  \\
  \hline
 FRCN& 07 & 74.5 &78.3  & 69.2 & 53.2 &36.6  &77.3  &78.2  &82.0  &40.7  &72.7  &67.9  &79.6  &79.2  &73.0  &69.0  &30.1  &65.4  & 70.2 &75.8&65.8&66.9\\
  \hline
FRCN & 07+12 & 77.0 &78.1  &69.3  &59.4  &38.3  &81.6  &78.6  &86.7  &42.8  &78.8  &68.9  &84.7  &82.0  &76.6  &69.9  &31.8  & 70.1 & 74.8 &80.4&70.4&70.0\\
  \hline
Ours &  07& 79.5  & 79.4 & 73.3  &57.3  &54.8  & 77.4  &84.3  &80.5  & 48.0 & 78.8 &63.6  & 80.7 & 78.3 &79.2  &67.6  & 42.8 & 72.9 & 67.7 &76.0&76.9& 71.0\\
  \hline
Ours &  07+12 & 80.5 &79.9  &76.9  &59.5  & 56.1 &78.9& 83.3 & 81.5 & 52.8 & 83.3 & 69.0 & 84.6 & 80.8& 79.3& 68.5& 49.3& 71.2& 74.5& 78.2& 78.8& 73.3\\
  \hline
\end{tabular}}
\vspace{0.1in}
\caption{Object detection mAP (\%) on the VOC 2007 test dataset. Training with only VOC07 trainval data is denoted as 07, and training with V0C07 trainval and V0C12 trainval is denoted as 07+12. RCNN and FRCN results come from \cite{girshick2015fast}.}
\label{table:voc07_det}
\vspace{-1pc}
\end{table*}

\subsubsection{Clustering window-object relationship}
\label{sec:investigation_on_location_specific_labels}
Window-object relationship is clustered in our approach as introduced in Section \ref{sec:locationPretrain}.
Its effectiveness is evaluated in this section.  Steps \ref{pipeline:step1}, \ref{pipeline:step2}, and \ref{pipeline:step4} are used.
The cropped image has only one setting of rotation and scale, i.e. $(r, \lambda)=(0^\circ, 1.2)$, which is the standard setting used in \cite{girshick2014rich, girshick2015fast}.  If only steps \ref{pipeline:step1} and \ref{pipeline:step4} are used, this corresponds to the RCNN baseline.
If  window-object relationship clustering is not used, the relative location and object class labels are used for learning features, which is the scheme in Fast RCNN \cite{girshick2015fast}, the mAP improvement is $0.2\%$. With clustering, the mAP improvement is $1.2\%$. Step \ref{pipeline:step2} is less effective without clustering and only brings $0.2\%$ improvement alone. Without clustering, a single regressor is learned for each class, relative locations and sizes cannot be accurately predicted and the learned features are less effective. Under each cluster, the configurations of locations and sizes are much simplified.

\subsubsection{Investigation on using multiple scales}
\label{sec:investigation_scales}

Based on the training pipeline using steps \ref{pipeline:step1}+\ref{pipeline:step2}+\ref{pipeline:step4} with window-object relationship clustering, Table \ref{table:scale} shows the influence of using multiple scales. The network with four scales has mAP $45.5\%$,  obtaining $4.2\%$ improvement compared with single scale. Based on the scale 1.2, the mAP improvements brought by an extra scale in descending order are 2.7, 1.8, and 0.8. This fits commonsense: a larger contextual region is more helpful in eliminating visual similarity between candidate boxes of different categories. More scales provide better performance, which shows that feature representations learned with different scales are complementary to each other. 

To figure out the effectiveness of employing multiple contextual scales for feature learning, we also run the configuration in which network parameters for all the four scales are shared and fixed to be that trained in scale 1.2. When using one shared network learned from scale 1.2, the employment of multiple contextual scales simply adds more visual cues, while training different networks for different scales, multiple contextual scales help feature learning through predicting window-object relationship which is our motivation. Compared with the network with shared network parameters, the networks with distinct parameters for different scales obtain $3.4\%$ mAP improvement. This shows that the use of multiple contextual scales is helpful to learn better features.

\subsubsection{Investigation on rotation}


Table \ref{table:rotation} shows the experimental results on using multiple rotation degrees and scales.
Table \ref{table:rotation} demonstrates that the performance improves mAP by $2.0\%$ for single scale and $0.3\%$ for multiple scales with the help of rotation.

\subsection{Overall results}
Ouyang \textit{et al.} \cite{ouyang2014deepid} showed that pre-training CNN with bounding boxes of objects instead of whole images in step \ref{pipeline:step1} could improve the detection accuracy significantly. It is also well known that using the bounding box regression \cite{girshick2014rich} to refine the locations of candidate windows in the last step of the detection pipeline is effective. In order to compete with the state-of-the-art, we incorporate the two existing technologies into our framework to boost the performance in the final evaluation.

Table \ref{table:ilsvrc14_det} summarizes the top ranked results on val2 and test datasets from ILSVRC2014 object challenge and demonstrates the effectiveness of our training pipeline. Flair \cite{vandeSandeCVPR2014} was the winner of ILSCRC2013. GoogleNet, DeepID-Net, DeepInsight, UvA-Euvision and Berkeley Vision were the top-ranked participants of ILSVRC2014 and GoogleNet was the winner.


Table \ref{table:voc07_det} reports the results on PASCAL VOC. Since the state-of-art approach Fast RCNN (FRCN) \cite{girshick2015fast} reported their performance of models trained on both VOC07 trainval and VOC12 trainval, we also evaluate our approach with the same training strategy. It has significant improvement on sate-of-the-art. It also outperforms the approaches of directly predicting bounding box locations from images.

\section{Conclusion}
This paper proposes a training pipeline that uses the window-object relationship for improving the representation learning. In order to help the CNN to estimate these relationships, multiple scales of contextual informations and rotations are utilized. Extensive component-wise experimental evaluation on ILSVRC14 object detection dataset validate the improvement from the proposed training pipeline. Our approach outperforms the sate-of-the-art on both ILSVRC14 and PASCAL VOC07 datasets.

{\small
\bibliographystyle{ieee}
\bibliography{egbib}
}

\end{document}